\documentclass[10pt,twocolumn,letterpaper]{article}

\usepackage{cvpr}
\usepackage{times}
\usepackage{epsfig}
\usepackage{graphicx}
\usepackage{amsmath}
\usepackage{amssymb}
\usepackage{caption}
\usepackage{multirow}
\usepackage{boldline}

\usepackage[breaklinks=true,bookmarks=false]{hyperref}

\cvprfinalcopy 

\def\cvprPaperID{3285} 

\ifcvprfinal\fi
\begin{document}

\title{Occlusion Aware Unsupervised Learning of Optical Flow}

\author{Yang Wang$^1$ \hspace{0.2cm} Yi Yang$^1$ \hspace{0.2cm} Zhenheng Yang$^2$  \hspace{0.2cm} Liang Zhao$^1$ \hspace{0.2cm} Peng Wang$^1$ \hspace{0.2cm} Wei Xu$^{1,3}$\\
$^1$Baidu Research \hspace{0.2cm} $^2$ University of Southern California \\
$^3$National Engineering Laboratory for Deep Learning Technology and Applications\\
{\tt\small \{wangyang59, yangyi05, zhaoliang07, wangpeng54, wei.xu\}@baidu.com  zhenheny@usc.edu}
}


\maketitle

\begin{abstract}
It has been recently shown that a convolutional neural network can learn optical flow estimation with unsupervised learning. However, the performance of the unsupervised methods still has a relatively large gap compared to its supervised counterpart. Occlusion and large motion are some of the major factors that limit the current unsupervised learning of optical flow methods. In this work we introduce a new method which models occlusion explicitly and a new warping way that facilitates the learning of large motion. Our method shows promising results on Flying Chairs, MPI-Sintel and KITTI benchmark datasets. Especially on KITTI dataset where abundant unlabeled samples exist, our unsupervised method outperforms its counterpart trained with supervised learning.
\end{abstract}

\begin{figure*}
\begin{center}
    \centering
    \includegraphics[width=.85\textwidth]{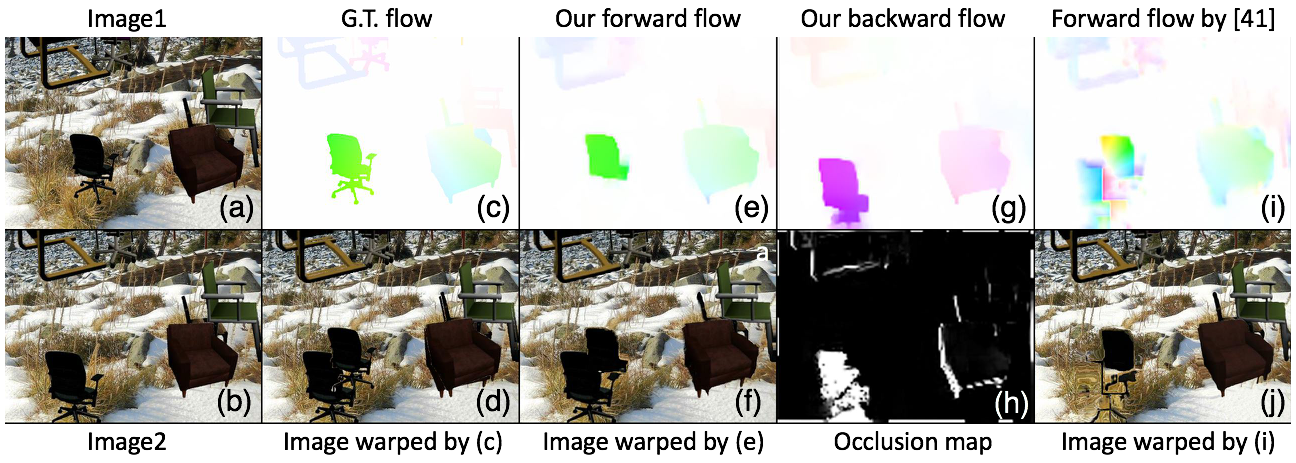}
    \captionof{figure}{(a) Input frame 1. (b) Input frame 2. (c) Ground-truth optical flow. (d) Image warped by ground-truth optical flow. (e) Forward optical flow estimated by our method. (f) Image warped by our forward optical flow. (g) Backward optical flow estimated by our method. (h) Occlusion map for input frame 1 estimated by our backward optical flow. (i) Optical flow from ~\cite{ren2017unsupervised}. (j) Image warped by~\cite{ren2017unsupervised}.}
    \label{fig:chair}
\end{center}
\end{figure*}

\section{Introduction}
Video motion prediction, or namely optical flow, is a fundamental problem in computer vision. With the accurate optical flow prediction, one could estimate the 3D structure of a scene~\cite{forsyth2011computer}, segment moving objects based on motion cues~\cite{pathak2016learning}, track objects in a complicated environment~\cite{bouguet2001pyramidal}, and build important visual cues for many high level vision tasks such as video action recognition~\cite{simonyan2014two} and video object detection~\cite{zhu2017flow}.

Traditionally, optical flow is formulated as a variational optimization problem with the goal of finding pixel correspondences between two consecutive video frames~\cite{horn1981determining}.
With the recent development of deep convolutional neural networks (CNNs)~\cite{krizhevsky2012imagenet}, deep learning based methods have been adopted to learn optical flow estimation, where the networks are either trained to compute discriminative image features for patch matching~\cite{guney2016deep} or directly output the dense flow fields in an end-to-end manner~\cite{dosovitskiy2015flownet}.
One major advantage of the deep learning based methods compared to classical energy-based methods is the computational speed, where most state-of-the-art energy-based methods require 1-50 minutes to process a pair of images, while deep nets only need less than 100 milliseconds with a modern GPU.

Since most deep networks are built to predict flow using two consecutive frames and trained with supervised learning~\cite{ilg2016flownet}, it would require a large amount of training data to obtain reasonably high accuracy~\cite{mayer2016large}.
Unfortunately, most large-scale flow datasets are from synthetic movies and ground-truth motion labels in real world videos are generally hard to annotate~\cite{Janai2017CVPR}.
To overcome this problem, unsupervised learning framework is proposed to utilize the resources of unlabeled videos ~\cite{jason2016back}.
The overall strategy behind those unsupervised methods is that instead of directly training the neural nets with ground-truth flow, they use a photometric loss that measures the difference between the target image and the (inversely) warped subsequent image based on the dense flow field predicted from the fully convolutional networks.
This allows the networks to be trained end-to-end with a large amount of unlabeled image pairs, overcoming the limitation from the lack of ground-truth flow annotations.

However, the performance of the unsupervised methods still has a relatively large gap compared to their supervised counterparts~\cite{ren2017unsupervised}.
To further improve unsupervised flow estimation, we realize that occlusion and large motion are among the major factors that limit the current unsupervised learning methods.
In this paper, we propose a new end-to-end deep neural architecture that carefully addresses these issues.

More specifically, the original baseline networks estimate motion and attempt to reconstruct every pixel in the target image. During reconstruction, there will be a fraction of pixels in the target image that have no source pixels due to occlusion. If we do not address this issue, it could limit the optical flow estimation accuracy since the loss function would prefer to compensate the occluded regions by moving other pixels. For example, in Fig.~\ref{fig:chair}, we would like to estimate the optical flow from frame 1 to frame 2, and reconstruct frame 1 by warping frame 2 with the estimated flow. Let us focus on the chair in the bottom left corner of the image. It moves in the down-left direction, and some part of the background is occluded by it. When we warp frame 2 back to frame 1 using the ground-truth flow (Fig.~\ref{fig:chair}c), the resulting image (Fig.~\ref{fig:chair}d) has two chairs in it. The chair on the top-right is the real chair, while the chair on the bottom-left is due to the occluded part of the background. Because the ground-truth flow of the background is zero, the chair in frame 2 is carried back to frame 1 to fill in the occluded background. Therefore, frame 2 warped by the ground-truth optical flow does not fully reconstruct frame 1. From the other perspective, if we use photometric loss of the entire image to guide the unsupervised learning of optical flow, the occluded area would not get the correct flow, which is illustrated in Fig.~\ref{fig:chair}i. It has an extra chair in the flow trying to fill the occluded background with nearby pixels of similar appearance, and the corresponding warped image Fig.~\ref{fig:chair}j has only one chair in it. 

To address this issue,  we explicitly allow the network to exploit the occlusion prediction caused by motion and incorporate it into the loss function. More concretely, we estimate the backward optical flow (Fig.~\ref{fig:chair}g) and use it to generate the occlusion map for the warped frame (Fig.~\ref{fig:chair}h). The white area in the occlusion map denotes the area in frame 1 that does not have a correspondence in frame 2. We train the network to only reconstruct the non-occluded area and do not penalize differences in the occluded area, so that the image warped by our estimated forward optical flow (Fig.~\ref{fig:chair}e) can have two chairs in it (Fig.~\ref{fig:chair}f) without incurring extra loss for the network.      

Our work differs from previous unsupervised learning methods in four aspects. 1) We proposed a new end-to-end neural network that handles occlusion. 2) We developed a new warping method that can facilitate unsupervised learning of large motion. 3) We further improved the previous FlowNetS by introducing extra warped inputs during the decoder phase. 4) We introduced histogram equalization and channel representation that are useful for optical flow estimation. The last three components are created to mainly tackle the issue of large motion estimation.

As a result, our method significantly improves the unsupervised learning based optical flow estimation on multiple benchmark dataset including Flying Chairs, MPI-Sintel and KITTI. Our unsupervised networks even outperforms its supervised counterpart~\cite{dosovitskiy2015flownet} on KITTI benchmark, where labeled data is limited compared to unlabeled data.


\section{Related Work}

Optical flow has been intensively studied in the past few decades~\cite{horn1981determining,lucas1981iterative,black1996robust,sun2010secrets,menze2015discrete}. Due to page limitation, we will briefly review the classical approaches and the recent deep learning approaches.

\textbf{Optical flow estimation.}
Optical flow estimation was introduced as a fundamental computer vision problem since the pioneering works~\cite{horn1981determining,lucas1981iterative}. 
Starting from then, the accuracy of optical flow estimation has
been improving steadily as evidenced by the results on Middlebury~\cite{baker2011database} and MPI-Sintel~\cite{butler2012naturalistic} benchmark dataset.
Most classical optical flow algorithms belong to the variants of the energy minimization problem with the brightness constancy and spatial smoothness assumptions~\cite{brox2004high,revaud2015epicflow}.
Other trends include a coarse-to-fine estimation or a hierarchical framework to deal with large motion~\cite{brox2011large,weinzaepfel2013deepflow,chen2013large,bailer2015flow}, a design of loss penalty to improve the robustness to lighting change and motion blur~\cite{zabih1994non,stein2004efficient,hafner2013census,vogel2013evaluation}, and a more sophisticated framework to handle occlusion~\cite{alvarez2007symmetrical,sun2010layered} which we will describe in more details in the next subsection.


\textbf{Occlusion-aware optical flow estimation.}
Since occlusion is a consequence of depth and motion, it is inevitable to model occlusion in order to accurately estimate flow.
Most existing methods jointly estimate optical flow and occlusion. 
Based on the methodology, we divide them into three major groups.
The first group treats occlusion as outliers and predict target pixels in the occluded regions as a constant value or through interpolation~\cite{strecha2004probabilistic,ayvaci2010occlusion,ayvaci2012sparse,unger2012joint}.
The second group deals with occlusion by exploiting the symmetric property of optical flow and ignoring the loss penalty on predicted occluded regions ~\cite{sun2005symmetric,alvarez2007symmetrical,hur2017mirrorflow}.
The last group builds more sophisticated frameworks such as modeling depth or a layered representation of objects to reason about occlusion~\cite{sun2010layered,sun2014local,yamaguchi2014efficient,sevilla2016optical}.
Our model is similar to the second group, such that we do not take account the difference where the occlusion happens into the loss function. To the best of our knowledge, we are the first to incorporate such kind of method with a neural network in an end-to-end trainable fashion. 
This helps our model to obtain more robust flow estimation around the occlusion boundary~\cite{ince2008occlusion,ballester2012tv}.

\textbf{Deep learning for optical flow.}
The success of deep learning innovates new optical flow models.
\cite{guney2016deep} uses deep nets to extract discriminative features to compute optical flow through patch matching.
\cite{bai2016exploiting} further extends the patch matching based methods by adding additional semantic information.
Later, \cite{bailer2017cnn} proposes a robust thresholded hinge loss for Siamese networks to learn CNN-based patch matching features.
\cite{xu2017accurate} accelerates the processing of patch matching cost volume and obtains optical flow results with high accuracy and fast speed.

Meanwhile, \cite{dosovitskiy2015flownet,ilg2016flownet} propose FlowNet to directly compute dense flow prediction on every pixel through fully convolutional neural networks and train the networks with end-to-end supervised learning.
\cite{ranjan2016optical} demonstrates that with a spatial pyramid network predicting in a coarse-to-fine fashion, a simple and small network can work quite accurately and efficiently on flow estimation.
Later, \cite{hur2016joint} proposes a method for jointly estimating optical flow and temporally consistent semantic segmentation with CNN.
The deep learning based methods obtain competitive accuracy across many benchmark optical flow datasets including MPI-Sintel~\cite{xu2017accurate} and KITTI~\cite{ilg2016flownet} with a relatively faster computational speed.
However, the supervised learning framework limits the extensibility of these works due to the lack of ground-truth flow annotation in other video datasets.

\textbf{Unsupervised learning for optical flow.}
\cite{patraucean2015spatio} first introduces an end-to-end differentiable neural architecture that allows unsupervised learning for video motion prediction and reports preliminary results on a weakly-supervised semantic segmentation task.
Later, \cite{jason2016back,ren2017unsupervised,ahmadi2016unsupervised} adopt a similar unsupervised learning architecture with a more detailed performance study on multiple optical flow benchmark datasets.
A common philosophy behind these methods is that instead of directly supervising with ground-truth flow, these methods utilize the Spatial Transformer Networks~\cite{jaderberg2015spatial} to warp the current images to produce a target image prediction and use photometric loss to guide back-propagation~\cite{finn2016unsupervised}.
The whole framework can be further extended to estimate the depth, camera motion and optical flow simultaneously in an end-to-end manner~\cite{vijayanarasimhan2017sfm}.
This overcomes the flow annotation problem, but the flow estimation accuracy in previous works still lags behind the supervised learning methods.
In this paper, we show that unsupervised learning can obtain competitive results to supervised learning models. After the initial submission of this paper, we became aware of a concurrent work~\cite{Meister:2018:UUL} which tries to solve the occlusion problem in unsupervised optical flow learning with a symmetric-based approach.  

\begin{figure}
\begin{center}
   \includegraphics[width=0.99\linewidth]{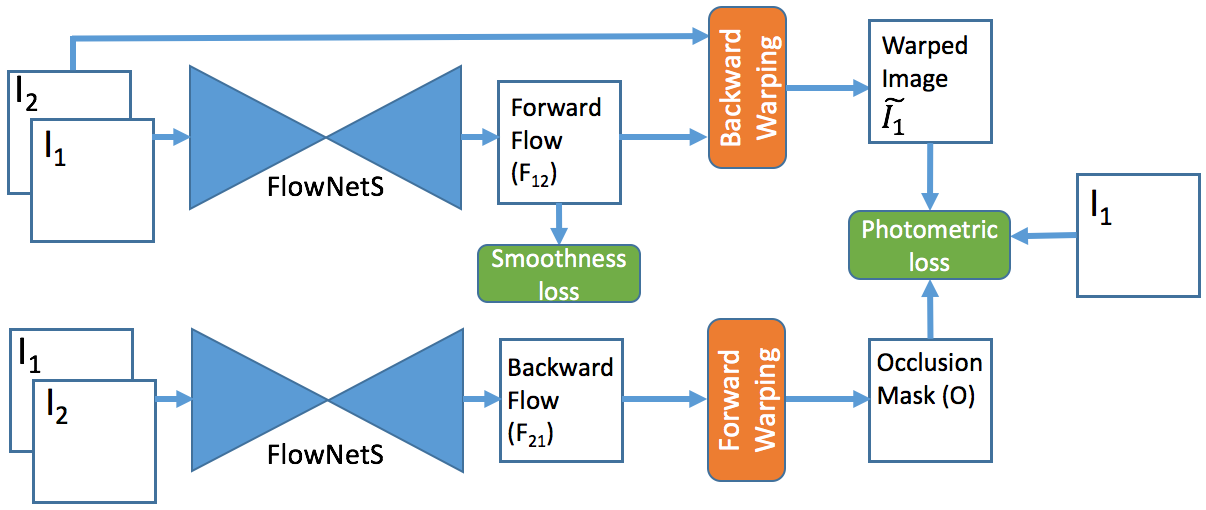}
\end{center}
   \caption{Our network architecture. It contains two copies of FlowNetS\cite{dosovitskiy2015flownet} with shared parameters which estimates forward and backward optical flow respectively. The forward warping module generates an occlusion map from the backward flow. The backward  warping module generates the warped image that is used to compare against the original frame 1 over the non-occluded area. There is also a smoothness term applied to the forward optical flow.}
\label{fig:network}
\end{figure}

\section{Network Structure and Method}
We first give an overview of our network structure and then describe each of its components in details. 

\textbf{Overall structure.} The schematic structure of our neural network is depicted in Fig.~\ref{fig:network}. Our network contains two copies of FlowNetS with shared parameters. The upper FlowNetS takes two stacked images ($I_{1}$ and $I_{2}$) as input and outputs the forward optical flow ($F_{12}$) from $I_{1}$ to $I_{2}$. The lower FlowNetS takes the reverse stacked images ($I_{2}$ and $I_{1}$) as input and outputs the backward flow ($F_{21}$) from $I_2$ to $I_1$. 

The forward flow $F_{12}$ is used to warp $I_2$ to reconstruct $\widetilde{I}_1$ through a Spatial Transformer Network similar to \cite{jason2016back}. We call this {\em backward warping}, since the warping direction is different from the flow direction. The backward flow $F_{21}$ is used to generate the occlusion map ($O$) by {\em forward warping}. The occlusion map indicates the region in $I_1$ that is correspondingly occluded in $I_2$ (\ie region in $I_1$ that does not have a correspondence in $I_2$). 

The loss for training our network contains two parts: a photometric term ($L_p$) and a smoothness term ($L_s$).
For the photometric term, we compare the warped image $\widetilde{I}_1$ and the original target image $I_1$ {\em in the non-occluded region} to obtain the photometric loss $L_p$. 
Note that this is a key difference between our method and previous unsupervised learning methods.
We also add a smoothness loss $L_s$ applied to $F_{12}$ to encourage a smooth flow solution.

\textbf{Forward warping and occlusion map.} We model the non-occluded region in $I_1$ as the range of $F_{21}$~\cite{alvarez2007symmetrical}, which can be calculated with the following equation,
\begin{equation*}
\begin{split}
V(x, y) = \sum_{i=1}^{W}\sum_{j=1}^{H}&\max\left(0, 1-|x - (i+F_{21}^{x}(i, j))|\right) \\
& \cdot \max\left(0, 1-|y - (j+F_{21}^{y}(i, j))|\right)
\end{split}
\end{equation*}
where $V(x, y)$ is the {\em range map} value at location $(x, y)$. $(W, H)$ are the image width and height, and $(F_{21}^{x}, F_{21}^{y})$ are the horizontal and vertical components of $F_{21}$.

\begin{figure}
\begin{center}
   \includegraphics[width=0.9\linewidth]{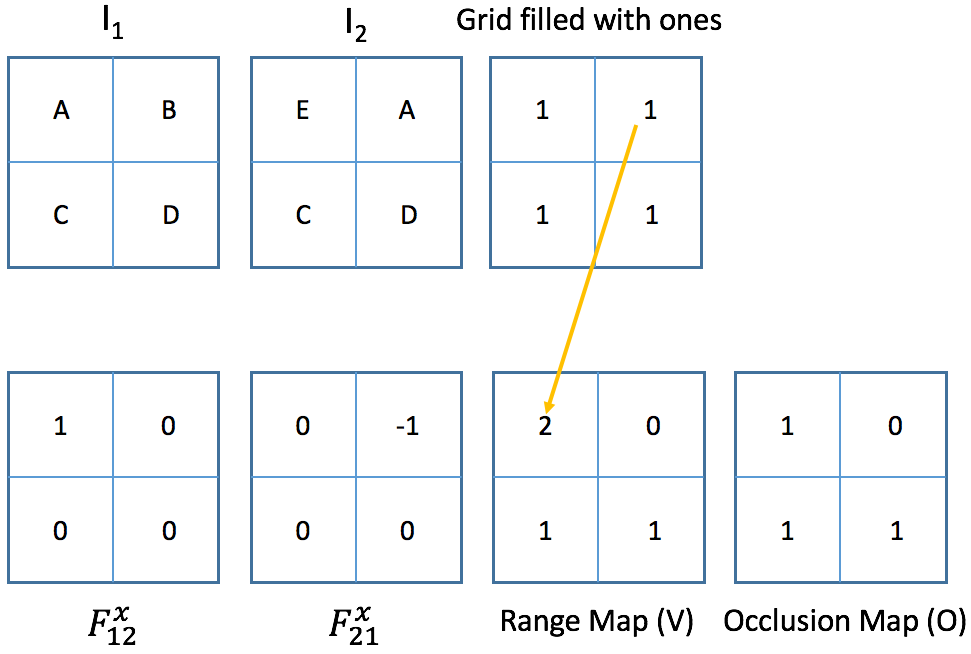}
\end{center}
   \caption{Illustration of the forward warping module demonstrating how the {\em occlusion map} is generated using the backward optical flow. Here we only have horizontal component optical flow $F^x_{12}$ and $F^x_{21}$ where 1 denotes moving right, -1 denote moving left and 0 denotes stationary. In the occlusion map, 0 denotes occluded and 1 denotes non-occluded.}
\label{fig:occlusion}
\end{figure}

Since $F_{21}$ is continuous, the location of a pixel after being translated by a floating number might not be exactly on an image grid. We use reversed bilinear sampling to distribute the weight of the translated pixel to its nearest neighbors. The {\em occlusion map} $O$ can be obtained by simply thresholding the range map $V$ at the value of 1 and results in a soft map with value between 0 and 1. $O(x, y) = \min(1, V(x, y))$. The whole forward warping module is differentiable and can be trained end-to-end with the rest of the network. 

In order to better illustrate the forward warping module, we provide a toy example in Fig.~\ref{fig:occlusion}. $I_1$ and $I_2$ have only 4 pixels each, in which different letters represent different pixel values. The flow and reversed flow only have horizontal component which we show as $F^x_{12}$ and $F^x_{21}$. The motion from $I_1$ to $I_2$ is that pixel A moves to the position of B and covers it, while pixel E in the background appears in $I_2$. To calculate the occlusion map, we first create an image filled with ones and then translate them according to $F_{21}$. Therefore, the one at the top-right corner is translated to the top-left corner leaving the top-right corner at the value of zero. The top-right corner (B) of $I_1$ is occluded by pixel A and can not find its corresponding pixel in $I_2$ which is consistent with the formulation we discussed above. 

\begin{figure}
\begin{center}
   \includegraphics[width=0.8\linewidth]{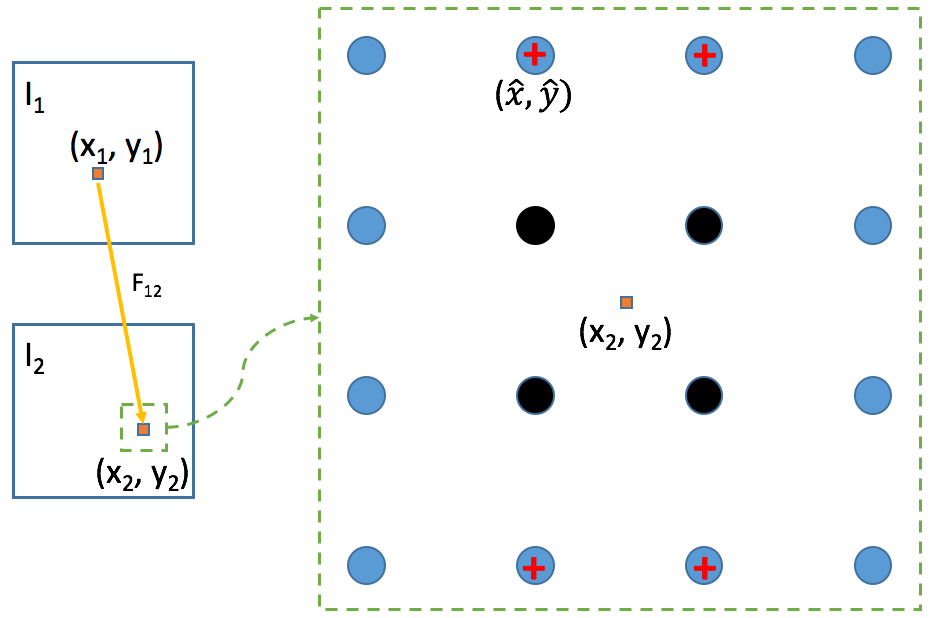}
\end{center}
   \caption{Illustration of the backward warping module with an enlarged search space. The large green box on the right side is a zoom view of the small green box on the left side.}
\label{fig:warp}
\end{figure}

\textbf{Backward warping with a larger search space.} The backward warping module is used to reconstruct $\widetilde{I}_1$ from $I_2$ with forward optical flow $F_{12}$. The method adopted here is similar to \cite{jason2016back,ren2017unsupervised} except that we include a larger search space. The problem with the original warping method is that the warped pixel value only depends on its four nearest neighbors, so if the target position is far away from the proposed position, the network will not get meaningful gradient signals. For example in Fig.~\ref{fig:warp}, a particular pixel lands in the position of $(x_2, y_2)$ proposed by the estimated optical flow, and its value is a weighted sum of its four nearest neighbors. However, if the true optical flow land the pixel at $(\hat{x}, \hat{y})$, the network would not learn the correct gradient direction, and thus stuck at a local minimum. This problem is particularly severe in the case of large motion. Although one could use a multi-scale image pyramid to tackle the large motion problem, if the moving object is small or has a similar color to the background, the motion might not be visible in small scale images.

More concretely, when we use the estimated optical flow $F_{12}$ to warp $I_2$ back to reconstruct $\widetilde{I}_1$ at a grid point $(x_1, y_1)$, we first translate the grid point $(x_1, y_1)$ in $I_1$ (the yellow square) to $(x_2, y_2) = (x_1+F_{12}^{x}(x_1, y_1), y_1+F_{12}^{y}(x_1, y_1))$ in $I_2$. Because the point $(x_2, y_2)$ is not on the grid point in $I_2$, we need to do bilinear sampling to obtain its value. Normally, the value at $(x_2, y_2)$ is a weighted sum of its four nearest neighbors (black dots in the zoomed view on the right side of Fig.~\ref{fig:warp}). We instead first search an enlarged neighbor (e.g. the blue dots at the outer circle in Fig.~\ref{fig:warp} together with the four nearest neightbors) around the point $(x_2, y_2)$. For instance, if in the enlarged neighbor of point $(x_2, y_2)$,  the point that has the closest value to the target value $I_1(x_1, y_1)$  is $(\hat{x}, \hat{y})$, we assign the value at the point $(x_2, y_2)$ to be a weighted sum of values at $(\hat{x}, \hat{y})$ and three other symmetrical points (points labeled with red crosses in Fig.~\ref{fig:warp}) with respect to point $(x_2, y_2)$. By doing this, we can provide the neural network with gradient pointing towards the location of $(\hat{x}, \hat{y})$.

\textbf{Loss term.} The loss of our network contains two components: a photometric loss ($L_p$) and a smoothness loss ($L_s$). We compute the photometric loss using the Charbonnier penalty formula $\Psi(s) = \sqrt{s^2 + 0.001^2}$ over the non-occluded regions with both image brightness and image gradient. 
\small
\begin{equation*}
L_p^1 = \big[\sum\limits_{i, j} \Psi(\widetilde{I}_1(i, j) - I_1(i, j)) \cdot O(i, j)\big] / \big[\sum\limits_{i, j} O(i, j) \big]
\end{equation*}
\begin{equation*}
L_p^2 = \big[\sum\limits_{i, j} \Psi(\nabla \widetilde{I}_1(i, j) - \nabla I_1(i, j)) \cdot O(i, j)\big] / \big[\sum\limits_{i, j} O(i, j)\big]
\end{equation*}
\normalsize
where $O$ is the occlusion map defined in the above section, and $i$, $j$ together indexes over pixel coordinates. The loss is normalized by the total non-occluded area size to prevent trivial solutions.  

For the smoothness loss, we adopt an edge-aware formulation similar to~\cite{godard2016unsupervised}, because motion boundaries usually coincide with image boundaries. Since the occluded area does not have a photometric loss, the optical flow estimation in the occluded area is solely guided by the smoothness loss. By using an edge-aware smoothness penalty, the optical flow in the occluded area would be similar to the values in its neighbor that has the closest appearance. We use both first-order and second-order derivatives of the optical flow in the smoothness loss term.
\begin{equation*}
\begin{split}
L_s^1 = & \sum_{i, j} \sum_{d \in x, y} \Psi\left(|\partial_d F_{12}(i, j)| e^{-\alpha|\partial_d I_1(i, j)|}\right) 
\end{split}
\end{equation*}
\begin{equation*}
\begin{split}
L_s^2 = & \sum_{i, j} \sum_{d \in x, y} \Psi\left(|\partial_d^2 F_{12}(i, j)| e^{-\alpha|\partial_d I_1(i, j)|}\right) 
\end{split}
\end{equation*}
where $\alpha$ controls the weight of edges, and $d$ indexes over partial derivative on $x$ and $y$ directions. 
The final loss is a weighted sum of the above four terms, \begin{equation*}
L = \gamma_1 L_p^1 + \gamma_2 L_p^2 + \gamma_3 L_s^1 + \gamma_4 L_s^2
\end{equation*}

\begin{figure}
\begin{center}
   \includegraphics[width=0.99\linewidth]{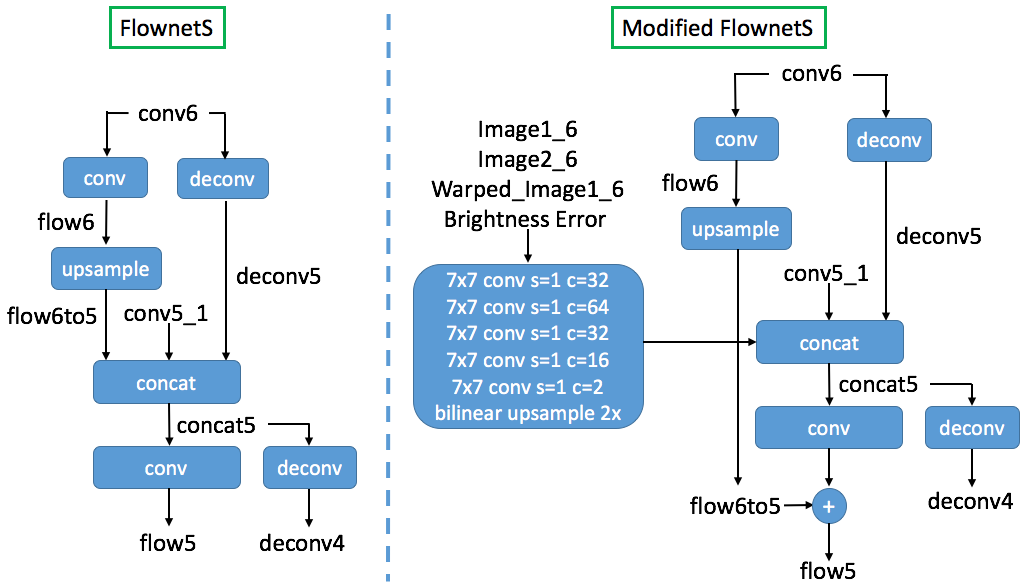}
\end{center}
   \caption{Our modification to the FlowNetS structure at one of the decoding stage. On the left, we show the original FlowNetS structure. On the right, we show our modification of the FlowNetS structure.  conv6 and conv5\_1 are features extracted in the encoding phase and named after \cite{dosovitskiy2015flownet}. Image1\_6 and Image2\_6 are input images downsampled 64 times. The decoding stages at other scales are modified accordingly.}
\label{fig:flownet}
\end{figure}

\textbf{Flow network details.} Our inner flow network is adopted from FlowNetS~\cite{dosovitskiy2015flownet}. Same as FlowNetS, we use a multi-scale scheme to guide the unsupervised learning by down-sampling images to different smaller scales.  The only modification we made to the FlowNetS structure is that from coarser to finer scale during the refinement phase, we add the image warped by the coarser optical flow estimation and its corresponding photometric error map as extra inputs to estimate the finer scale optical flow in a fashion similar to FlowNet2~\cite{ilg2016flownet}. By doing this, each layer only needs to estimate the residual between the coarse and fine scale. The detailed network structure can be found in Fig.~\ref{fig:flownet}. Our modification only increases the number of parameters by 2\% compared to the original FlownetS, and it moderately improves the result as seen in the later ablation study.   

\textbf{Preprocessing.} In order to have better contrast for moving objects in the down-sampled images, we preprocess the image pairs by applying histogram equalization and augment the RGB image with a channel representation. The detailed channel representation can be found in~\cite{sevilla2014optical}. We find both preprocessing steps improve the final optical flow estimation results.

\begin{table*}[t]
\centering
\begin{tabular}{ c | l | c | c  c  c  c | c  c  c  c }
   \hlineB{3}
   & Methods & Chairs & \multicolumn{2}{c}{Sintel Clean} & \multicolumn{2}{c|}{Sintel Final} & \multicolumn{2}{c}{KITTI 2012} & \multicolumn{2}{c}{KITTI 2015} \\
                   &  & test  & train & test & train & test & train & test & train & test \\
   \hline
   \multirow{6}{*}{\rotatebox[origin=c]{90}{Supervise}} &
   FlowNetS~\cite{dosovitskiy2015flownet}    & 2.71 & 4.50 & 7.42 & 5.45 & 8.43 & 8.26 & -- & -- & --  \\
   & FlowNetS+ft~\cite{dosovitskiy2015flownet} & -- & (3.66) & 6.96 & (4.44) & 7.76 & 7.52 & 9.1 & -- & -- \\
   & SpyNet~\cite{ranjan2016optical}      & \bf{2.63} & 4.12 & 6.69 & 5.57 & 8.43 & 9.12 & -- & -- & -- \\
   & SpyNet+ft~\cite{ranjan2016optical}   & --      & (3.17) & 6.64 & (4.32) & 8.36 & 8.25 & 10.1 & -- & -- \\
   & FlowNet2~\cite{ilg2016flownet}    & --      & \bf{2.02} & \bf{3.96} & \bf{3.14} & 6.02 & \bf{4.09} & -- & \bf{10.06} & -- \\
   & FlowNet2+ft~\cite{ilg2016flownet} & --      & (1.45) & 4.16 & (2.01) & \bf{5.74} & (1.28) & \bf{1.8} & (2.3) & \bf{11.48\%} \\
   \hline
   \multirow{6}{*}{\rotatebox[origin=c]{90}{Unsupervise}} &
   DSTFlow~\cite{ren2017unsupervised}    & 5.11 &  6.93 & 10.40 & 7.82 & 11.11 & 16.98 & -- & 24.30 & -- \\
   & DSTFlow-best~\cite{ren2017unsupervised} & 5.11 & (6.16) & 10.41 & (6.81) & 11.27 & 10.43 & 12.4 & 16.79 & 39\% \\
   & BackToBasic~\cite{jason2016back} & 5.3 & -- & -- & -- & -- & 11.3 & 9.9 & -- & -- \\
   & Ours           & \bf{3.30} & \bf{5.23} & 8.02 & \bf{6.34} & \bf{9.08} & 12.95 & -- & 21.30 & -- \\
   & Ours+ft-Sintel & 3.76   & (4.03) & \bf{7.95} & (5.95) & 9.15 & 12.9 & -- & 22.6 & --\\
   & Ours-KITTI & -- & 7.41 & -- & 7.92 & -- & \bf{3.55} & \bf{4.2} & \bf{8.88} & \bf{31.2\%} \\
  \hlineB{3}
\end{tabular}
\caption{Quantitative evaluation of our method on different benchmarks. The numbers reported here are all average end-point-error (EPE) except for the last column (KITTI2015 test) which is the percentage of erroneous pixels (Fl-all). A pixel is considered to be correctly estimated if the flow end-point error is \textless3px or \textless5\%. The upper part of the table contains supervised methods and lower part of the table contains unsupervised methods. For all metrics, smaller is better. The best number for each category is highlighted in bold. The numbers in parentheses are results from network trained on the same set of data, and hence are not directly comparable to other results.}
\label{tab:result}
\end{table*}

\section{Experimental Results}

We evaluate our methods on standard optical flow benchmark datasets including Flying Chairs~\cite{dosovitskiy2015flownet}, MPI-Sintel~\cite{butler2012naturalistic} and KITTI~\cite{geiger2012we}, and compare our results to existing deep learning based optical flow estimation (both supervised and unsupervised methods).
We use the standard endpoint error (EPE) measure as the evaluation metric, which is the average Euclidean distance between the predicted flow and the ground truth flow over all pixels.

\subsection{Implementation Details}

Our network is trained end-to-end using Adam optimizer~\cite{kingma2014adam} with $\beta_1 = 0.9$ and $\beta_2 = 0.999$. The learning rate is set to be $10^{-4}$ for training from scratch and $10^{-5}$ for fine-tuning. The experiments are performed on two Titan Z GPUs with a batch size of 8 or 16 depending on the input image resolution. The training converges after roughly a day. During training, we first assign equal weights to loss from different image scales and then progressively increase the weight on the larger scale image in a way similar to~\cite{mayer2016large}. The hyper-parameters $(\gamma_1, \gamma_2, \gamma_3, \gamma_4, \alpha)$ are set to be (1.0, 1.0, 10.0, 0.0, 10.0) for Flying Chairs and MPI-Sintel datasets, and (0.03, 3.0, 0.0, 10.0, 10.0) for KITTI dataset.Here we used higher weights of image gradient photometric loss and second-order smoothness loss for KITTI because the data has more lightning changes and its optical flow has more continuously varying intrinsic structure. In terms of data augmentaion, we only used horizontal flipping, vertical flipping and image pair order switching. During testing, our network only predicts forward flow, the total computational time on a Flying Chairs image pair is roughly 90 milliseconds with our Titian Z GPUs. Adding an extra 8 milliseconds for histogram equalization (an OpenCV CPU implementation), the total prediction time is around 100 milliseconds.

\subsection{Quantitative and Qualitative Results}
\begin{figure*}
\begin{center}
    \includegraphics[width=.8\textwidth]{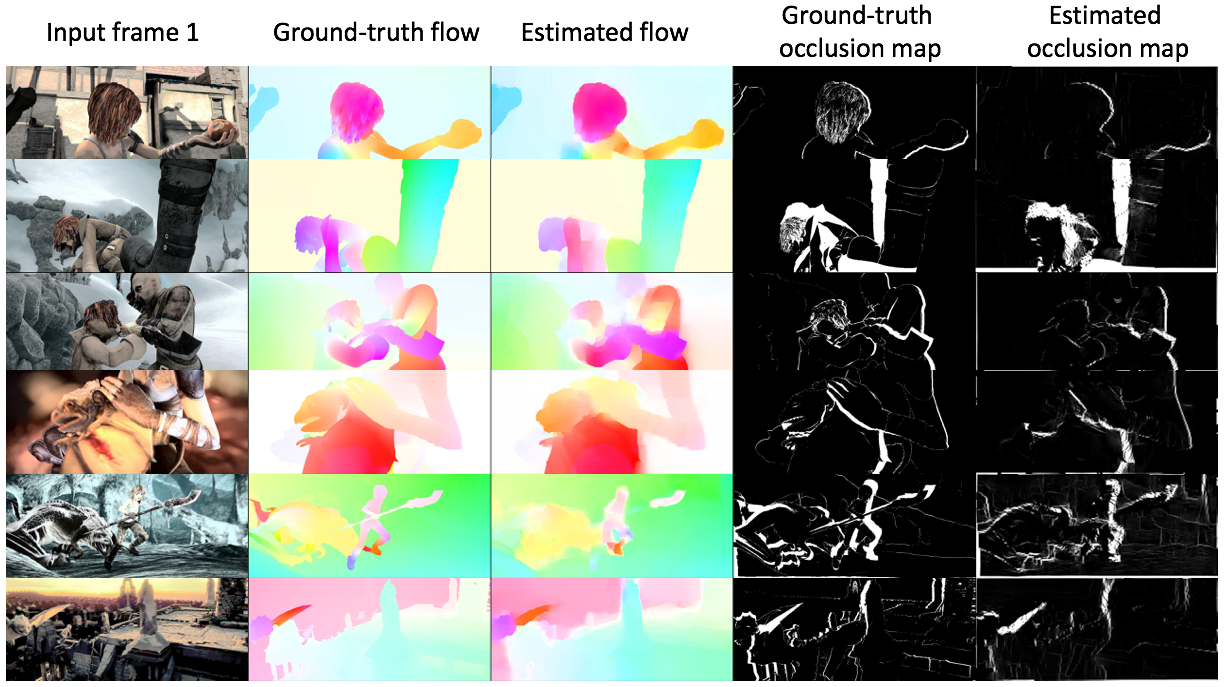}
    \captionof{figure}{Qualitative examples for Sintel dataset. The top three rows are from Sintel Clean and the bottom three rows are from Sintel Final.}
    \label{fig:sintel}
\end{center}
\end{figure*}

\begin{figure*}
\begin{center}
   \includegraphics[width=0.8\linewidth]{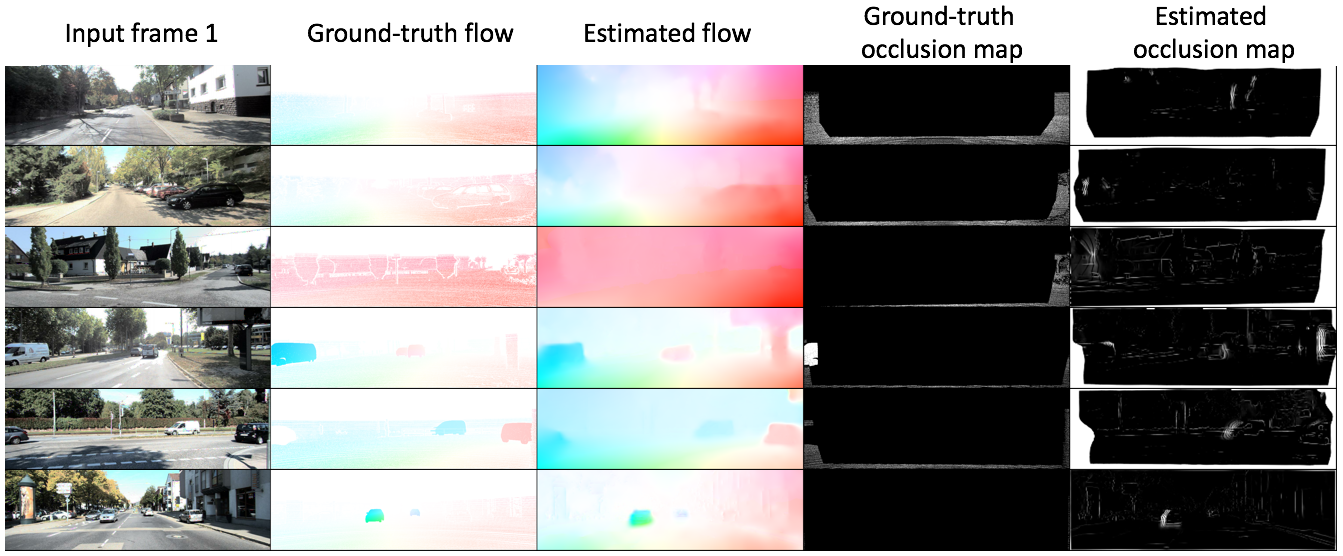}
   \caption{Qualitative examples for KITTI dataset. The top three rows are from KITTI 2012 and the bottom three rows are from KITTI 2015.}
\label{fig:kitti}
\end{center}
\end{figure*}

\begin{table*}[t]
\centering
\begin{tabular}{ c  c c c | c c  c }
   \hlineB{3}
   occlusion  & enlarged  &  modified & contrast  & Chairs  & Sintel Clean & Sintel Final \\
    handling  & search & FlowNet & enhancement & test & train & train \\
   \hline
                   &                    &                     &                    & 5.11 & 6.93 & 7.82 \\
   \hline
   \checkmark &            	  &                      &                    & 4.51 & 6.80 & 7.32 \\
   \checkmark & \checkmark &                    &                     & 4.27 & 6.49 & 7.11 \\
   \checkmark & \checkmark & \checkmark &                     & 4.14 & 6.38 & 7.08 \\
   		  &                        & \checkmark &                    & 4.62 & 6.60 & 7.33 \\
                   &                        & \checkmark & \checkmark & 4.04 & 6.09 & 7.04 \\
   \checkmark &                    & \checkmark & \checkmark  & 3.76 & 5.70 & 6.54 \\ 

   \checkmark & \checkmark & \checkmark & \checkmark & \bf{3.30} & \bf{5.23} & \bf{6.34} \\
  \hlineB{3}  
\end{tabular}
\caption{Ablation study}
\label{tab:ablation}
\end{table*}

Table~\ref{tab:result} summarizes the EPE of our method and previous state-of-the-art deep learning methods, including FlowNet~\cite{dosovitskiy2015flownet}, SpyNet~\cite{ranjan2016optical}, FlowNet2~\cite{ilg2016flownet}, DSTFlow~\cite{ren2017unsupervised} and BackToBasic~\cite{jason2016back}. Because DSTFlow reported multiple variations of their results, we cite their best number across all of their results in "DSTFlow-best" here. 

\textbf{Flying Chairs.} Flying Chairs is a synthetic dataset created by superimposing images of chairs on background images from Flickr. It was originally created for training FlowNet in a supervised manner~\cite{dosovitskiy2015flownet}. We use it to train our network without using any ground-truth flow. We randomly split the dataset into 95\% training and 5\% testing. We label this model as "Ours" in Table~\ref{tab:result}. Our EPE is significantly smaller than the previous unsupervised methods (\ie EPE decreases from 5.11 to 3.30) and is approaching the level of its corresponding supervised learning result (2.71). 

\textbf{MPI-Sintel.} Since MPI-Sintel is relatively small and only contains around a thousand image pairs, we use the training data from both clean and final pass (without ground-truth) to fine-tune our network pretrained on Flying Chairs and the resulting model is labeled as "Ours+ft-Sintel". Compared to other unsupervised methods, we achieve a much better performance (e.g., EPE decreases from 10.40 to 7.95 on Sintel Clean test). Note that fine-tuning did not improve much here, largely due to the small number of training data. Fig.\ref{fig:sintel} illustrates the qualitative result of our method on MPI-Sintel. 

\textbf{KITTI.} The KITTI dataset is recorded under real-world driving conditions, and it has more unlabeled data than labeled data. Unsupervised learning methods would have an advantage in this scenario since they can learn from the large amount of unlabeled data. The training data we use here is similar to \cite{ren2017unsupervised} which consists of multi-view extensions (20 frames for each sequence) from both KITTI2012 and KITTI2015. During training, we exclude two neighboring frames from the image pairs with ground-truth flow and testing pairs to avoid mixing training and testing data (\ie not including frame number 9-12 in each multi-view sequence). We train the model from scratch since the optical flow in KITTI dataset has its own domain spatial structure (different from Flying Chairs) and abundant data. We label this model as "Ours-KITTI" in Table~\ref{tab:result}. 

Table~\ref{tab:result} suggests that our method not only significantly outperforms existing unsupervised learning methods (\ie improves EPE from 9.9 to 4.2 on KITTI 2012 test), but also outperforms its supervised counterpart (FlowNetS+ft) by a large margin, although there is still a gap compared to the state-of-the-art supervised network FlowNet2. Fig.~\ref{fig:kitti} illustrates the qualitative results on KITTI. Our model correctly captures the occluded area caused by moving out of the frame. Our flow results are also free from the artifacts seen in DSTFlow (see \cite{ren2017unsupervised} Figure 4c) in the occlusion area.

\textbf{Occlusion Estimation.} We also evaluate our occlusion estimation on MPI-Sintel and KITTI dataset which provide ground-truth occlusion labels between two consecutive frames. Among the literatures, we only find limited reports on occlusion estimation accuracy. Table~\ref{tab:occlusion} shows the occlusion estimation performance by calculating the maximum F-measure introduced in~\cite{leordeanu2013locally}. On MPI-Sintel, our method has a comparable result with previous non-neural-network based methods~\cite{leordeanu2013locally,xu2012motion}. On KITTI we obtain 0.95 and 0.88 for KITTI2012 and KITTI2015 respectively (we did not find published occlusion estimation result on KITTI). Note that S2D used ground-truth occlusion maps to do supervised training of their occlusion model.  

\begin{table}
\centering
\begin{tabular}{ c | c c c  c }
   \hlineB{3}
    Method & Sintel & Sintel & KITTI & KITTI  \\
           & Clean  & Final  & 2012  & 2015 \\
    \hline
    Our       & 0.54  &  0.48 & 0.95 & 0.88 \\
   S2D~\cite{leordeanu2013locally}       &  --     &  \bf{0.57} &    --   &   --      \\
   MODOF~\cite{xu2012motion} &   --    & 0.48  &    --   &    --    \\
  \hlineB{3}  
\end{tabular}
\caption{Occlusion estimation evaluation. The numbers we present here is maximum F-measure. The S2D method is trained with ground-truth occlusion labels.} 
\label{tab:occlusion}
\end{table}

\subsection{Ablation Study}
We conduct systematic ablation analysis on different components added in our method. Table~\ref{tab:ablation} shows the overall effects of them on Flying Chairs and MPI-Sintel. Our starting network is a FlowNetS without occlusion handling, which is the same configuration as~\cite{ren2017unsupervised}. 

\textbf{Occlusion handling.} The top two rows in Table~\ref{tab:ablation} suggest that by only adding occlusion handling to the baseline network, the model improves its EPE from 5.11 to 4.51 on Flying-Chairs and from 7.82 to 7.32 on MPI-Sintel Final, which is significant. 

\textbf{Enlarged search.} The effect of enlarged search is also significant. The bottom two rows in Table~\ref{tab:ablation} show that adding enlarged search, the final EPE improves from 3.76 to 3.30 on Flying-Chairs and from 6.54 to 6.34 on MPI-Sintel Final. 

\textbf{Modified FlowNet.} A small modification to the FlowNet also improves significantly, as suggested in the 5-th row in Table~\ref{tab:ablation}. By only adding a 2\% more parameters and computation, the EPE improves from 5.11 to 4.62 on Flying-Chairs and from 7.82 to 7.33 on MPI-Sintel Final.

\textbf{Contrast enhancement.} We find that contrast enhancement is also a simple but very effective preprocessing step to improve the unsupervised optical flow learning. By comparing the 4th row and last row in Table~\ref{tab:ablation}, we find the final EPE improves from 4.14 to 3.30 on Flying-Chairs and 7.08 to 6.34 on MPI-Sintel Final.

\textbf{Combining all components.} We also find that sometimes one component is not significant by itself, but the overall model improves dramatically when we add all the 4 components into our framework.

\textbf{Effect of data.} We have tried to use more data from KITTI raw videos (60,000 samples compared to 25,000 samples used in the paper) to train our model, but we did not find any improvement. We have also tried to adopt the network structure from SpyNet~\cite{ranjan2016optical} and train them using our unsupervised method. However we did not get better result either, which suggests that the learning capability of our model is still the limiting factor, although we have pushed this forward by a large margin.

\section{Conclusion}
We present a new end-to-end unsupervised learning framework for optical flow prediction. We show that with modeling occlusion and large motion, our unsupervised approach yields competitive results on multiple benchmark datasets. 
This is promising since it opens a new path for training neural networks to predict optical flow with a vast amount of unlabeled videos and apply the flow estimation for more higher level computer vision tasks.

{\small
\bibliographystyle{ieee}
\bibliography{egbib}

\begin{thebibliography}{10}\itemsep=-1pt

\bibitem{ahmadi2016unsupervised}
A.~Ahmadi and I.~Patras.
\newblock Unsupervised convolutional neural networks for motion estimation.
\newblock In {\em Image Processing (ICIP), 2016 IEEE International Conference
  on}, pages 1629--1633. IEEE, 2016.

\bibitem{alvarez2007symmetrical}
L.~Alvarez, R.~Deriche, T.~Papadopoulo, and J.~S{\'a}nchez.
\newblock Symmetrical dense optical flow estimation with occlusions detection.
\newblock {\em International Journal of Computer Vision}, 75(3):371--385, 2007.

\bibitem{ayvaci2010occlusion}
A.~Ayvaci, M.~Raptis, and S.~Soatto.
\newblock Occlusion detection and motion estimation with convex optimization.
\newblock In {\em Advances in neural information processing systems}, pages
  100--108, 2010.

\bibitem{ayvaci2012sparse}
A.~Ayvaci, M.~Raptis, and S.~Soatto.
\newblock Sparse occlusion detection with optical flow.
\newblock {\em International Journal of Computer Vision}, 97(3):322--338, 2012.

\bibitem{bai2016exploiting}
M.~Bai, W.~Luo, K.~Kundu, and R.~Urtasun.
\newblock Exploiting semantic information and deep matching for optical flow.
\newblock In {\em European Conference on Computer Vision}, pages 154--170.
  Springer, 2016.

\bibitem{bailer2015flow}
C.~Bailer, B.~Taetz, and D.~Stricker.
\newblock Flow fields: Dense correspondence fields for highly accurate large
  displacement optical flow estimation.
\newblock In {\em Proceedings of the IEEE International Conference on Computer
  Vision}, pages 4015--4023, 2015.

\bibitem{bailer2017cnn}
C.~Bailer, K.~Varanasi, and D.~Stricker.
\newblock Cnn-based patch matching for optical flow with thresholded hinge
  embedding loss.
\newblock In {\em IEEE Conference on Computer Vision and Pattern Recognition
  (CVPR)}, 2017.

\bibitem{baker2011database}
S.~Baker, D.~Scharstein, J.~Lewis, S.~Roth, M.~J. Black, and R.~Szeliski.
\newblock A database and evaluation methodology for optical flow.
\newblock {\em International Journal of Computer Vision}, 92(1):1--31, 2011.

\bibitem{ballester2012tv}
C.~Ballester, L.~Garrido, V.~Lazcano, and V.~Caselles.
\newblock A tv-l1 optical flow method with occlusion detection.
\newblock {\em Pattern Recognition}, pages 31--40, 2012.

\bibitem{black1996robust}
M.~J. Black and P.~Anandan.
\newblock The robust estimation of multiple motions: Parametric and
  piecewise-smooth flow fields.
\newblock {\em Computer vision and image understanding}, 63(1):75--104, 1996.

\bibitem{bouguet2001pyramidal}
J.-Y. Bouguet.
\newblock Pyramidal implementation of the affine lucas kanade feature tracker
  description of the algorithm.
\newblock {\em Intel Corporation}, 5(1-10):4, 2001.

\bibitem{brox2004high}
T.~Brox, A.~Bruhn, N.~Papenberg, and J.~Weickert.
\newblock High accuracy optical flow estimation based on a theory for warping.
\newblock {\em Computer Vision-ECCV 2004}, pages 25--36, 2004.

\bibitem{brox2011large}
T.~Brox and J.~Malik.
\newblock Large displacement optical flow: descriptor matching in variational
  motion estimation.
\newblock {\em IEEE transactions on pattern analysis and machine intelligence},
  33(3):500--513, 2011.

\bibitem{butler2012naturalistic}
D.~J. Butler, J.~Wulff, G.~B. Stanley, and M.~J. Black.
\newblock A naturalistic open source movie for optical flow evaluation.
\newblock In {\em European Conference on Computer Vision}, pages 611--625.
  Springer, 2012.

\bibitem{chen2013large}
Z.~Chen, H.~Jin, Z.~Lin, S.~Cohen, and Y.~Wu.
\newblock Large displacement optical flow from nearest neighbor fields.
\newblock In {\em Proceedings of the IEEE Conference on Computer Vision and
  Pattern Recognition}, pages 2443--2450, 2013.

\bibitem{dosovitskiy2015flownet}
A.~Dosovitskiy, P.~Fischer, E.~Ilg, P.~Hausser, C.~Hazirbas, V.~Golkov,
  P.~van~der Smagt, D.~Cremers, and T.~Brox.
\newblock Flownet: Learning optical flow with convolutional networks.
\newblock In {\em Proceedings of the IEEE International Conference on Computer
  Vision}, pages 2758--2766, 2015.

\bibitem{finn2016unsupervised}
C.~Finn, I.~Goodfellow, and S.~Levine.
\newblock Unsupervised learning for physical interaction through video
  prediction.
\newblock In {\em Advances in Neural Information Processing Systems}, pages
  64--72, 2016.

\bibitem{forsyth2011computer}
D.~Forsyth and J.~Ponce.
\newblock {\em Computer vision: a modern approach}.
\newblock Upper Saddle River, NJ; London: Prentice Hall, 2011.

\bibitem{geiger2012we}
A.~Geiger, P.~Lenz, and R.~Urtasun.
\newblock Are we ready for autonomous driving? the kitti vision benchmark
  suite.
\newblock In {\em Computer Vision and Pattern Recognition (CVPR), 2012 IEEE
  Conference on}, pages 3354--3361. IEEE, 2012.

\bibitem{godard2016unsupervised}
C.~Godard, O.~Mac~Aodha, and G.~J. Brostow.
\newblock Unsupervised monocular depth estimation with left-right consistency.
\newblock In {\em CVPR}, volume~2, page~7, 2017.

\bibitem{guney2016deep}
F.~G{\"u}ney and A.~Geiger.
\newblock Deep discrete flow.
\newblock In {\em Asian Conference on Computer Vision}, pages 207--224.
  Springer, 2016.

\bibitem{hafner2013census}
D.~Hafner, O.~Demetz, and J.~Weickert.
\newblock Why is the census transform good for robust optic flow computation?
\newblock In {\em International Conference on Scale Space and Variational
  Methods in Computer Vision}, pages 210--221. Springer, 2013.

\bibitem{horn1981determining}
B.~K. Horn and B.~G. Schunck.
\newblock Determining optical flow.
\newblock {\em Artificial intelligence}, 17(1-3):185--203, 1981.

\bibitem{hur2016joint}
J.~Hur and S.~Roth.
\newblock Joint optical flow and temporally consistent semantic segmentation.
\newblock In {\em European Conference on Computer Vision}, pages 163--177.
  Springer, 2016.

\bibitem{hur2017mirrorflow}
J.~Hur and S.~Roth.
\newblock Mirrorflow: Exploiting symmetries in joint optical flow and occlusion
  estimation.
\newblock {\em arXiv preprint arXiv:1708.05355}, 2017.

\bibitem{ilg2016flownet}
E.~Ilg, N.~Mayer, T.~Saikia, M.~Keuper, A.~Dosovitskiy, and T.~Brox.
\newblock Flownet 2.0: Evolution of optical flow estimation with deep networks.
\newblock In {\em IEEE Conference on Computer Vision and Pattern Recognition
  (CVPR)}, volume~2, 2017.

\bibitem{ince2008occlusion}
S.~Ince and J.~Konrad.
\newblock Occlusion-aware optical flow estimation.
\newblock {\em IEEE Transactions on Image Processing}, 17(8):1443--1451, 2008.

\bibitem{jaderberg2015spatial}
M.~Jaderberg, K.~Simonyan, A.~Zisserman, et~al.
\newblock Spatial transformer networks.
\newblock In {\em Advances in Neural Information Processing Systems}, pages
  2017--2025, 2015.

\bibitem{Janai2017CVPR}
J.~Janai, F.~Güney, J.~Wulff, M.~Black, and A.~Geiger.
\newblock Slow flow: Exploiting high-speed cameras for accurate and diverse
  optical flow reference data.
\newblock In {\em Conference on Computer Vision and Pattern Recognition
  (CVPR)}, 2017.

\bibitem{jason2016back}
J.~Y. Jason, A.~W. Harley, and K.~G. Derpanis.
\newblock Back to basics: Unsupervised learning of optical flow via brightness
  constancy and motion smoothness.
\newblock In {\em Computer Vision--ECCV 2016 Workshops}, pages 3--10. Springer,
  2016.

\bibitem{kingma2014adam}
D.~Kingma and J.~Ba.
\newblock Adam: A method for stochastic optimization.
\newblock {\em arXiv preprint arXiv:1412.6980}, 2014.

\bibitem{krizhevsky2012imagenet}
A.~Krizhevsky, I.~Sutskever, and G.~E. Hinton.
\newblock Imagenet classification with deep convolutional neural networks.
\newblock In {\em Advances in neural information processing systems}, pages
  1097--1105, 2012.

\bibitem{leordeanu2013locally}
M.~Leordeanu, A.~Zanfir, and C.~Sminchisescu.
\newblock Locally affine sparse-to-dense matching for motion and occlusion
  estimation.
\newblock In {\em Proceedings of the IEEE International Conference on Computer
  Vision}, pages 1721--1728, 2013.

\bibitem{lucas1981iterative}
B.~D. Lucas, T.~Kanade, et~al.
\newblock An iterative image registration technique with an application to
  stereo vision.
\newblock 1981.

\bibitem{mayer2016large}
N.~Mayer, E.~Ilg, P.~Hausser, P.~Fischer, D.~Cremers, A.~Dosovitskiy, and
  T.~Brox.
\newblock A large dataset to train convolutional networks for disparity,
  optical flow, and scene flow estimation.
\newblock In {\em Proceedings of the IEEE Conference on Computer Vision and
  Pattern Recognition}, pages 4040--4048, 2016.

\bibitem{Meister:2018:UUL}
S.~Meister, J.~Hur, and S.~Roth.
\newblock {UnFlow}: Unsupervised learning of optical flow with a bidirectional
  census loss.
\newblock In {\em AAAI}, New Orleans, Louisiana, Feb. 2018.

\bibitem{menze2015discrete}
M.~Menze, C.~Heipke, and A.~Geiger.
\newblock Discrete optimization for optical flow.
\newblock In {\em German Conference on Pattern Recognition}, pages 16--28.
  Springer, 2015.

\bibitem{pathak2016learning}
D.~Pathak, R.~Girshick, P.~Doll{\'a}r, T.~Darrell, and B.~Hariharan.
\newblock Learning features by watching objects move.
\newblock In {\em Proc. CVPR}, volume~2, 2017.

\bibitem{patraucean2015spatio}
V.~Patraucean, A.~Handa, and R.~Cipolla.
\newblock Spatio-temporal video autoencoder with differentiable memory.
\newblock {\em arXiv preprint arXiv:1511.06309}, 2015.

\bibitem{ranjan2016optical}
A.~Ranjan and M.~J. Black.
\newblock Optical flow estimation using a spatial pyramid network.
\newblock In {\em IEEE Conference on Computer Vision and Pattern Recognition
  (CVPR)}, volume~2, 2017.

\bibitem{ren2017unsupervised}
Z.~Ren, J.~Yan, B.~Ni, B.~Liu, X.~Yang, and H.~Zha.
\newblock Unsupervised deep learning for optical flow estimation.
\newblock In {\em AAAI}, pages 1495--1501, 2017.

\bibitem{revaud2015epicflow}
J.~Revaud, P.~Weinzaepfel, Z.~Harchaoui, and C.~Schmid.
\newblock Epicflow: Edge-preserving interpolation of correspondences for
  optical flow.
\newblock In {\em Proceedings of the IEEE Conference on Computer Vision and
  Pattern Recognition}, pages 1164--1172, 2015.

\bibitem{sevilla2016optical}
L.~Sevilla-Lara, D.~Sun, V.~Jampani, and M.~J. Black.
\newblock Optical flow with semantic segmentation and localized layers.
\newblock In {\em Proceedings of the IEEE Conference on Computer Vision and
  Pattern Recognition}, pages 3889--3898, 2016.

\bibitem{sevilla2014optical}
L.~Sevilla-Lara, D.~Sun, E.~G. Learned-Miller, and M.~J. Black.
\newblock Optical flow estimation with channel constancy.
\newblock In {\em European Conference on Computer Vision}, pages 423--438.
  Springer, 2014.

\bibitem{simonyan2014two}
K.~Simonyan and A.~Zisserman.
\newblock Two-stream convolutional networks for action recognition in videos.
\newblock In {\em Advances in neural information processing systems}, pages
  568--576, 2014.

\bibitem{stein2004efficient}
F.~Stein.
\newblock Efficient computation of optical flow using the census transform.
\newblock In {\em DAGM-symposium}, volume 2004, pages 79--86. Springer, 2004.

\bibitem{strecha2004probabilistic}
C.~Strecha, R.~Fransens, and L.~J. Van~Gool.
\newblock A probabilistic approach to large displacement optical flow and
  occlusion detection.
\newblock In {\em ECCV Workshop SMVP}, pages 71--82. Springer, 2004.

\bibitem{sun2014local}
D.~Sun, C.~Liu, and H.~Pfister.
\newblock Local layering for joint motion estimation and occlusion detection.
\newblock In {\em Proceedings of the IEEE Conference on Computer Vision and
  Pattern Recognition}, pages 1098--1105, 2014.

\bibitem{sun2010secrets}
D.~Sun, S.~Roth, and M.~J. Black.
\newblock Secrets of optical flow estimation and their principles.
\newblock In {\em Computer Vision and Pattern Recognition (CVPR), 2010 IEEE
  Conference on}, pages 2432--2439. IEEE, 2010.

\bibitem{sun2010layered}
D.~Sun, E.~B. Sudderth, and M.~J. Black.
\newblock Layered image motion with explicit occlusions, temporal consistency,
  and depth ordering.
\newblock In {\em Advances in Neural Information Processing Systems}, pages
  2226--2234, 2010.

\bibitem{sun2005symmetric}
J.~Sun, Y.~Li, S.~B. Kang, and H.-Y. Shum.
\newblock Symmetric stereo matching for occlusion handling.
\newblock In {\em Computer Vision and Pattern Recognition, 2005. CVPR 2005.
  IEEE Computer Society Conference on}, volume~2, pages 399--406. IEEE, 2005.

\bibitem{unger2012joint}
M.~Unger, M.~Werlberger, T.~Pock, and H.~Bischof.
\newblock Joint motion estimation and segmentation of complex scenes with label
  costs and occlusion modeling.
\newblock In {\em Computer Vision and Pattern Recognition (CVPR), 2012 IEEE
  Conference on}, pages 1878--1885. IEEE, 2012.

\bibitem{vijayanarasimhan2017sfm}
S.~Vijayanarasimhan, S.~Ricco, C.~Schmid, R.~Sukthankar, and K.~Fragkiadaki.
\newblock Sfm-net: Learning of structure and motion from video.
\newblock {\em arXiv preprint arXiv:1704.07804}, 2017.

\bibitem{vogel2013evaluation}
C.~Vogel, S.~Roth, and K.~Schindler.
\newblock An evaluation of data costs for optical flow.
\newblock In {\em German Conference on Pattern Recognition}, pages 343--353.
  Springer, 2013.

\bibitem{weinzaepfel2013deepflow}
P.~Weinzaepfel, J.~Revaud, Z.~Harchaoui, and C.~Schmid.
\newblock Deepflow: Large displacement optical flow with deep matching.
\newblock In {\em Proceedings of the IEEE International Conference on Computer
  Vision}, pages 1385--1392, 2013.

\bibitem{xu2017accurate}
J.~Xu, R.~Ranftl, and V.~Koltun.
\newblock Accurate optical flow via direct cost volume processing.
\newblock In {\em The IEEE Conference on Computer Vision and Pattern
  Recognition (CVPR)}, July 2017.

\bibitem{xu2012motion}
L.~Xu, J.~Jia, and Y.~Matsushita.
\newblock Motion detail preserving optical flow estimation.
\newblock {\em IEEE Transactions on Pattern Analysis and Machine Intelligence},
  34(9):1744--1757, 2012.

\bibitem{yamaguchi2014efficient}
K.~Yamaguchi, D.~McAllester, and R.~Urtasun.
\newblock Efficient joint segmentation, occlusion labeling, stereo and flow
  estimation.
\newblock In {\em European Conference on Computer Vision}, pages 756--771.
  Springer, 2014.

\bibitem{zabih1994non}
R.~Zabih and J.~Woodfill.
\newblock Non-parametric local transforms for computing visual correspondence.
\newblock In {\em European conference on computer vision}, pages 151--158.
  Springer, 1994.

\bibitem{zhu2017flow}
X.~Zhu, Y.~Wang, J.~Dai, L.~Yuan, and Y.~Wei.
\newblock Flow-guided feature aggregation for video object detection.
\newblock {\em arXiv preprint arXiv:1703.10025}, 2017.

\end{thebibliography}
}

\end{document}